\pdfoutput=1
\documentclass[10pt,twocolumn,letterpaper]{article}
\usepackage{btas}
\usepackage{times}
\usepackage{epsfig}
\usepackage{graphicx}
\usepackage{epsfig} 
\usepackage{graphicx}
\usepackage{amsmath}
\usepackage{amssymb}
\usepackage{algorithm,algpseudocode}
\usepackage{amsmath}
\usepackage{amsfonts}
\usepackage{amssymb}
\usepackage{enumitem}
\usepackage{stfloats}
\usepackage{subcaption}
\usepackage{balance}
\usepackage{pgfplots}
\usetikzlibrary{external}

\pdfoutput=1

\floatname{algorithm}{Procedure}

\btasfinalcopy 

\ifbtasfinal\fi

\begin{document}
%
\title{Robust Morph-Detection at Automated Border Control Gate using Deep Decomposed 3D Shape \& Diffuse Reflectance}
%
%
%
\author{Jag Mohan Singh ~ Raghavendra Ramachandra ~ Kiran B. Raja ~ Christoph Busch  \\
\bf{Norwegian Biometrics Laboratory, NTNU, Norway} \\
\bf{\texttt{\{jag.m.singh; raghavendra.ramachandra; kiran.raja; christoph.busch\}@ntnu.no}} \\
}
\maketitle

\begin{abstract}
Face recognition is widely employed in Automated Border Control (ABC) gates, which verify the face image on passport or electronic Machine Readable Travel Document (eMTRD) against the captured image to confirm the identity of the passport holder. In this paper, we present a robust morph detection algorithm that is based on differential morph detection. The proposed method decomposes the bona fide image captured from the ABC gate and the digital face image extracted from the eMRTD into the diffuse reconstructed image and a quantized normal map. The extracted features are further used to learn a linear classifier (SVM) to detect a morphing attack based on the assessment of differences between the bona fide image from the ABC gate and the digital face image extracted from the passport. Owing to the availability of multiple cameras within an ABC gate, we extend the proposed method to fuse the classification scores to generate the final decision on morph-attack-detection. To validate our proposed algorithm, we create a morph attack database with overall 588 images, where bona fide are captured in an indoor lighting environment with a Canon DSLR Camera with one sample per subject and correspondingly images from ABC gates. We benchmark our proposed method with the existing state-of-the-art and can state that the new approach significantly outperforms previous approaches in the ABC gate scenario.
\end{abstract}


%

\section{Introduction}\label{intro}

\begin{figure}[t!]
\centering
\includegraphics[width=1.0\linewidth]{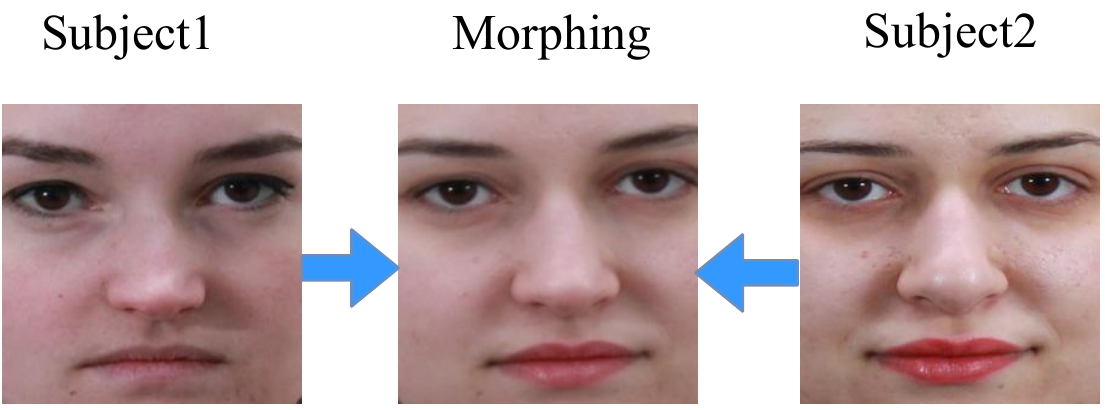} 
 \caption{Digital morphing example from our database}
 \label{fig:MorphingExample}
\end{figure}

Face recognition systems (FRS) are widely deployed at border crossings, which use Automated Border Control (ABC) gates. The deployment has ever increased since member states of the International Civil Aviation Organization (ICAO) follow ICAO's specification 9303 and store a standardized digital face image in the electronic Machine Readable Travel Document (eMRTD). However, FRS has shown to be vulnerable with respect to morphed face images - a new image as a result of a weighted linear combination of two input images, as shown in Figure~\ref{fig:MorphingExample}. The generated morphed image challenges the FRS as it can be used to verify two unique identities (individuals), defeating the FRS's ability to verify unique subjects~\cite{Raghu17}.
The challenge becomes severe as some countries issue the passport based on the digital photo uploaded by the applicant, which can provide an opportunity to upload a morphed image that can later be verified by an FRS~\cite{Raghu17,Ulrich18}. Several countermeasures have been proposed for Morphing Attack Detection (MAD). MAD can be broadly classified into No-Reference MAD (NR-MAD), which uses a single image for MAD and Differential MAD (D-MAD), which uses an image pair that includes a trusted live capture, and an image extracted from eMRTD. In addition, both MAD methods (NR-MAD and D-MAD) do or do not anticipate potential artifacts that have been introduced in the image signal with an optional print and scan process of the facial image \cite{Raghu19}.
In the rest of the paper, we present related work in Section~\ref{related}, our proposed algorithm in Section~\ref{proposedalgorithmsection}, followed by experimental setup, and results in Section~\ref{resultssection}, and conclusions and future-work in Section~\ref{conclusionsection}.

\section{Related Work}\label{related}
\begin{figure*}[h!]
\centering
\includegraphics[width=0.90\linewidth]{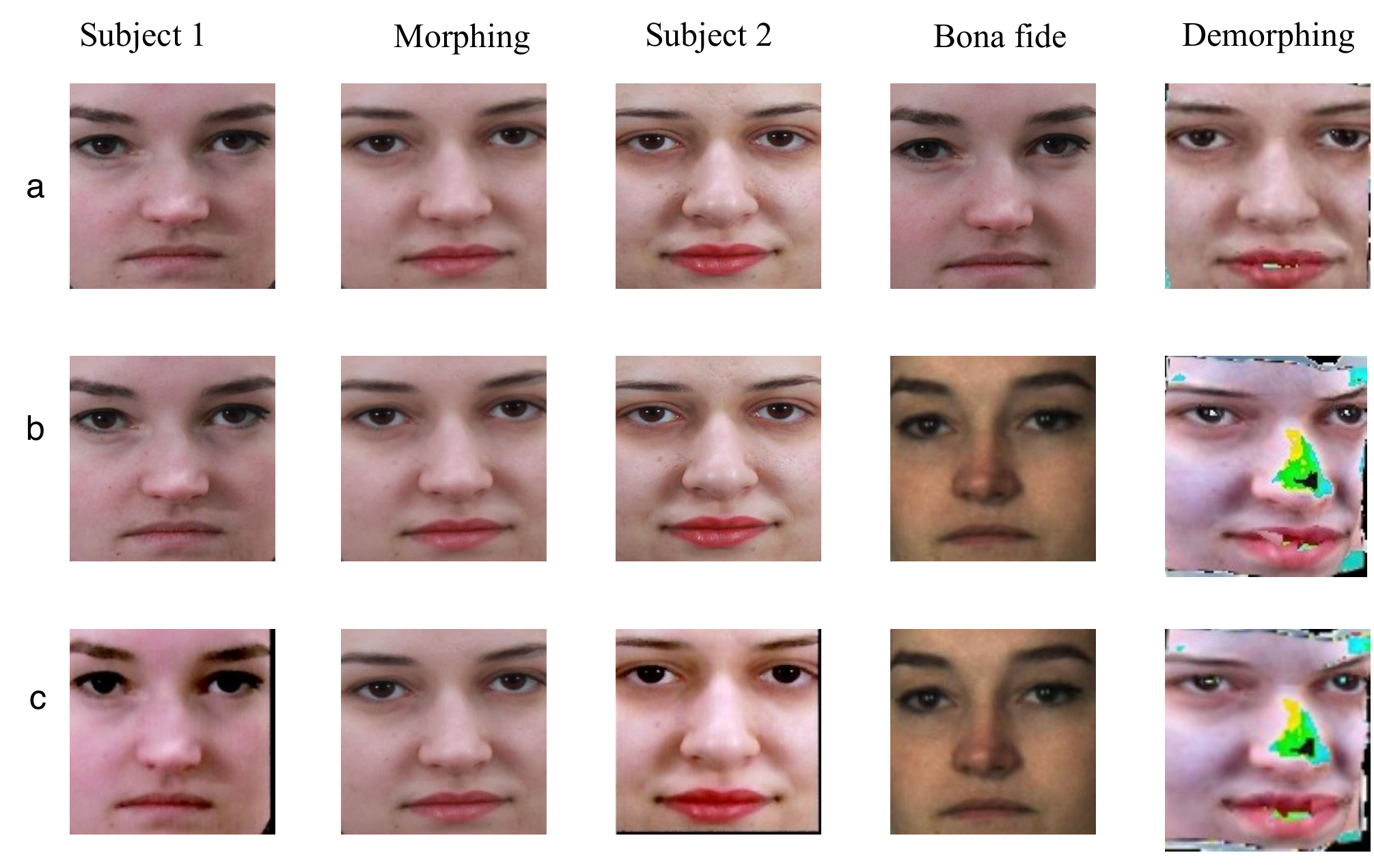} 
\caption{Demorphing using Image Subtraction based technqiue by Ferrara et. al \cite{Ferrara18} of subjects in different conditions fails especially in (b), and (c) where Bona fide Image is from our dataset. Rows: Digital Images in (a) Bona fide Image captured in similar lighting \& (b), Print-Scan (Inkjet EPSON \textsuperscript{TM}) Images (c). Results are based on our own implementation.}\label{fig:DemorphingComparison}
\end{figure*}

In this section, we review the related-work for D-MAD for which there are several algorithms, such as using landmark shifts proposed by Damer et. al~\cite{Naser18}, texture-descriptors based approach proposed by Scherhag et. al~\cite{Ulrich18}, and image subtraction based approach proposed by Ferrara et. al~\cite{Ferrara18}. 
The authors in~\cite{Naser18} conduct a face alignment using a common facial landmark detector ~\cite{King09} for each image and compute a distance-vector subsequently from landmark locations to train an SVM-RBF for morph detection. The authors in~\cite{Ulrich18} also employ the face-alignment from~\cite{King09}, followed by computing the vector differences between texture-descriptors such as LBP~\cite{LBPPaper}, BSIF~\cite{BSIFPaper}, or SIFT~\cite{SIFTPaper}. The vector difference is then used to train an SVM-RBF for differential morph detection. 
One of the existing state-of-the-art (SOTA) schemes presented by authors in~\cite{Ferrara18} tries to invert the morphing process using image subtraction. 
The authors observe that given the warping functions and alpha value, one could perfectly demorph a morphed image. However, in a practical scenario, the warping functions, and alpha value are unknown, so the authors obtain warping functions by face alignment, and prescribe $\alpha=0.45$ for best quality demorphing. 
The following are the limitations of current SOTA in differential MAD, landmark shifts could occur due to pose changes, texture-descriptor features would have reduced efficacy in the presence of lighting, pose, and print-scan artifacts~\cite{Raghu17}, and image subtraction methods would have reduced efficiency in the presence of lighting, pose, and print-scan artifacts as shown in Figure~\ref{fig:DemorphingComparison} some of which are also shown in~\cite{Ferrara18}.

\begin{figure*}[b]
\centering
\includegraphics[width=0.85\linewidth]{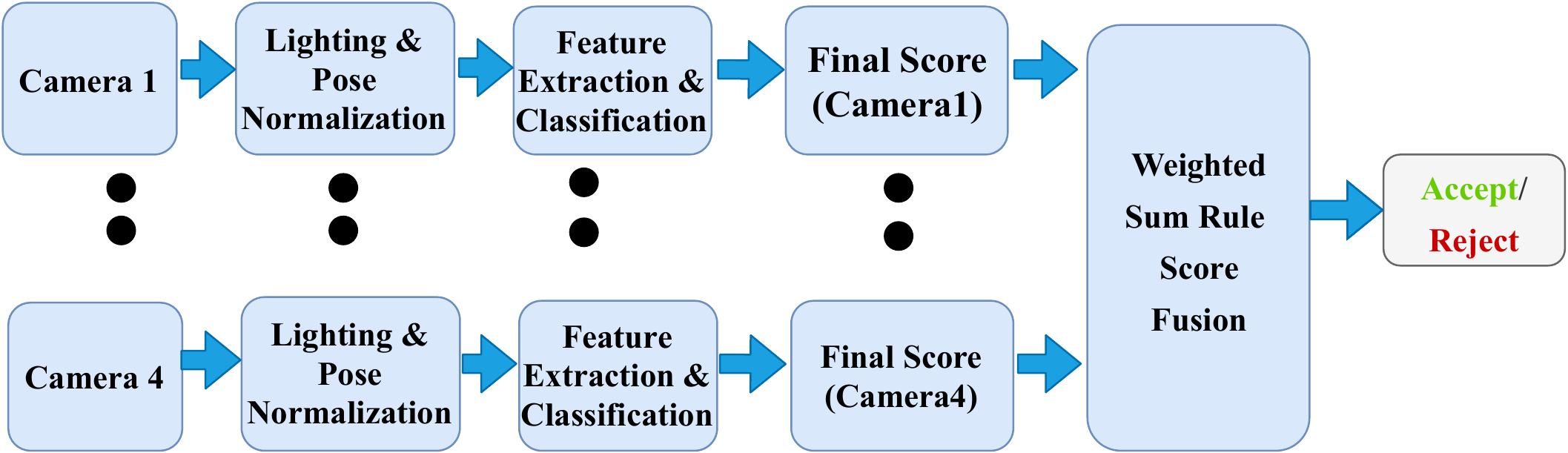} 
 \caption{Pipeline of our approach showing the fusion of scores from Camera1, Camera2, Camera3 and Camera4 where each camera features are in-turn generated by fusion.}\label{fig:ProposedAlgorithm}
\end{figure*}
In a real border control scenario, the subject is verified with the captured face image from the ABC gate, which is compared against the image stored in the eMRTD. This is what we modeled in our work. We leverage this to verify if the image on eMRTD is morphed by looking at the 3D shape and reflectance for both captured images from ABC and image within the eMRTD. Specifically, we look at the normal-map and the diffuse reconstructed image, to devise a classifier that can distinguish bona fide (non-morphed) images from morphed images. We assert that the morphed image presents significantly inconsistent information within the image as compared to the non-morphed image. It has further to be noted that many ABC gates operate with multiple cameras, which enable us to reinforce the decision with fusion approaches to detect a morphing attack in a better manner, as demonstrated in our work. To the best of our knowledge, this is the first method to explore the strengths of a multi-camera capture set-up in border control operations to detect the morphing attacks. To assert our approach, we create a new database with bona fide images of $39$ subjects in an ideal enrolment setting and correspondingly the probe images of the same $39$ subjects, which were captured while crossing the ABC gate. The images from the $39$ subjects are used to create morphed images ($90$). 

The key contributions of this work, therefore, can be summarized as:
\begin{itemize}
    \item Presents a new database of morphed images and trusted live capture probe images captured in a realistic border crossing scenario with ABC gates.
   \item Presents a new approach employing the inherent border crossing scenario to detect the morphing attacks using a fusion of scores from a quantized normal-map approach and diffuse reconstructed image characteristics.
  \item Presents an extensive evaluation of state-of-art D-MAD techniques to benchmark the proposed algorithm, and demonstrate the superiority of the proposed algorithm.
\end{itemize}

\begin{figure}[h!]
\centering
\includegraphics[width=0.85\linewidth]{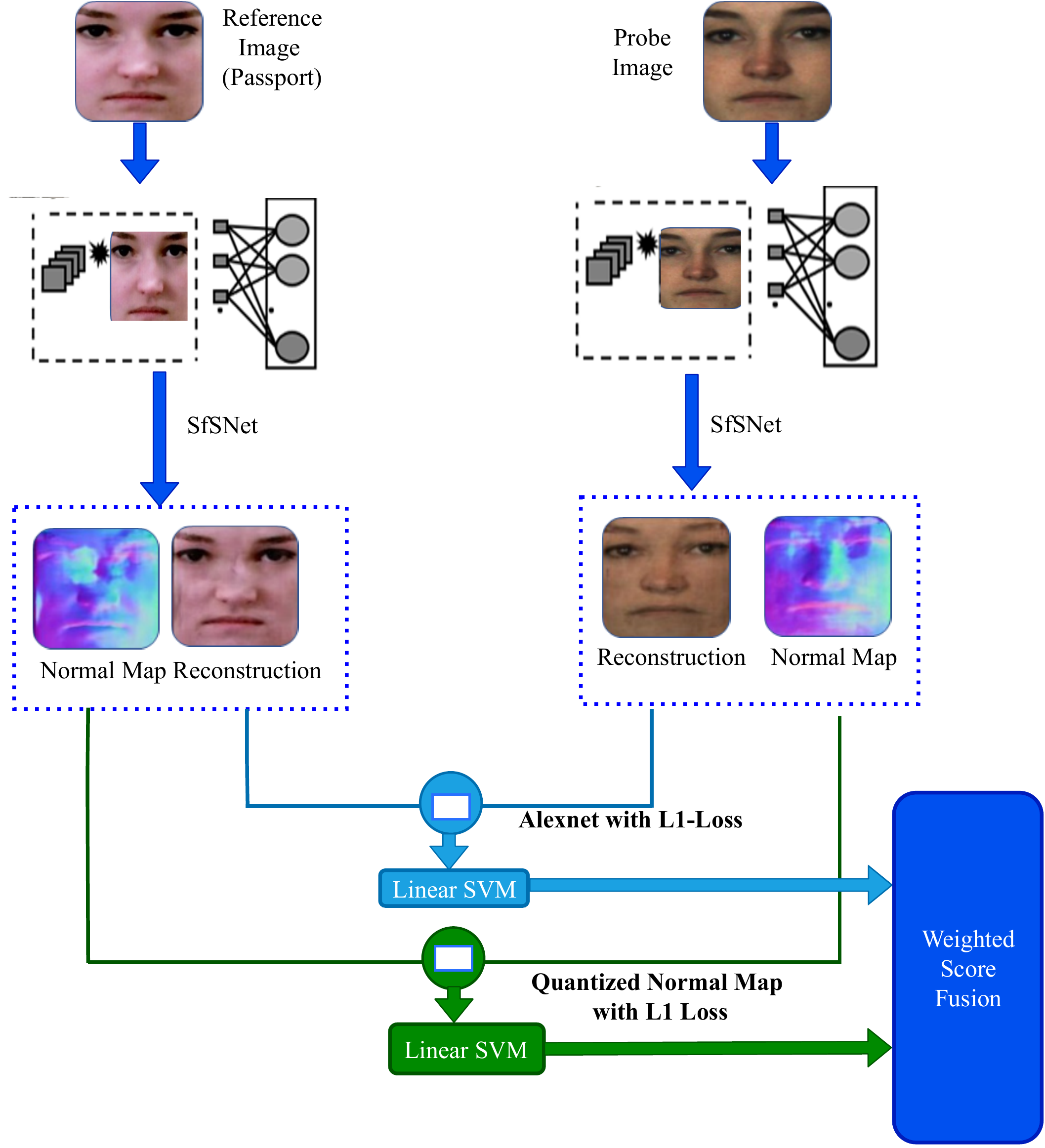} 
 \caption{Illustration of feature extraction and classification for each camera}\label{fig:FeatureExtractionAndClassification}
\end{figure}

\section{Proposed Algorithm}\label{proposedalgorithmsection}
In this section, we describe the proposed algorithm for robust morph detection at an ABC gate. In our approach, the probe face image, which is captured at the ABC gate, is compared with a face image from the eMRTD. The ABC gate face image and the digital face image from the eMRTD would likely have intensity changes due to lighting differences in the capture environments, pose changes due to the capture subject interaction, image quality differences along with the additional noises introduced in the print-scan process preceding the storing of a given digital face image in the eMRTD. Given that these changes may not optimally help in determining a morph attack, we formulate the problem of morphing attack detection first by normalizing the pose changes in the image, further to which we compute the features for D-MAD. The pipeline of the proposed approach is depicted in Figure~\ref{fig:ProposedAlgorithm}, where pose normalization is carried out first.
Further to this, we extract the features to learn a robust classifier, as shown in Figure~\ref{fig:FeatureExtractionAndClassification} for each camera. Given the availability of multiple cameras, we further propose a weighted sum-rule score level fusion for scores from each camera. Each of the components of the proposed method is further detailed, as provided in the subsequent sections. 

\subsection{Pose Normalization}\label{lightingposenormalization}
 We also do pose normalization using the method from authors in~\cite{King09} as the face images from ABC Gate could be in a non-frontal pose. The method we use for pose normalization is based on the key-points which are automatically detected in a face, and it makes the line joining the eye-centers horizontal.
\subsection{Feature Extraction and Classification}\label{featureExtractionClassification}

Given the images are now normalized for pose using the method described in Section~\ref{lightingposenormalization}, we proceed to extract the features. We, therefore, decompose an input image $I$ into diffuse reconstructed image $I(p)$ and a normal map $n(p)$, which represents the shape of the face. We choose SfSNet~\cite{Soumyadip18}, as it can decompose a single input image into the diffuse reconstructed image, normal-map, albedo-map $\rho$, and 2nd order spherical harmonic based lighting coefficients $l_{nm}$. 
The diffuse reconstructed image can be written as with second order spherical harmonics using~\cite{Basri03} as follows:
\begin{equation}\label{eq1}
I(p) = \rho r(n(p))
\end{equation}
where r(n(p)) which is reflectance of the material, is given by
\begin{equation}\label{eq2}
r(n(p)) = \sum_{n=0}^{n=2} \sum_{m=-n}^n l_{nm} r_{nm}(n(p))
\end{equation}
where $l_{nm}$ for $n=0$ are used from the ambient coefficients identified in Section~\ref{lightingposenormalization}.

\begin{figure}[h!]
\centering
\includegraphics[width=0.75\linewidth]{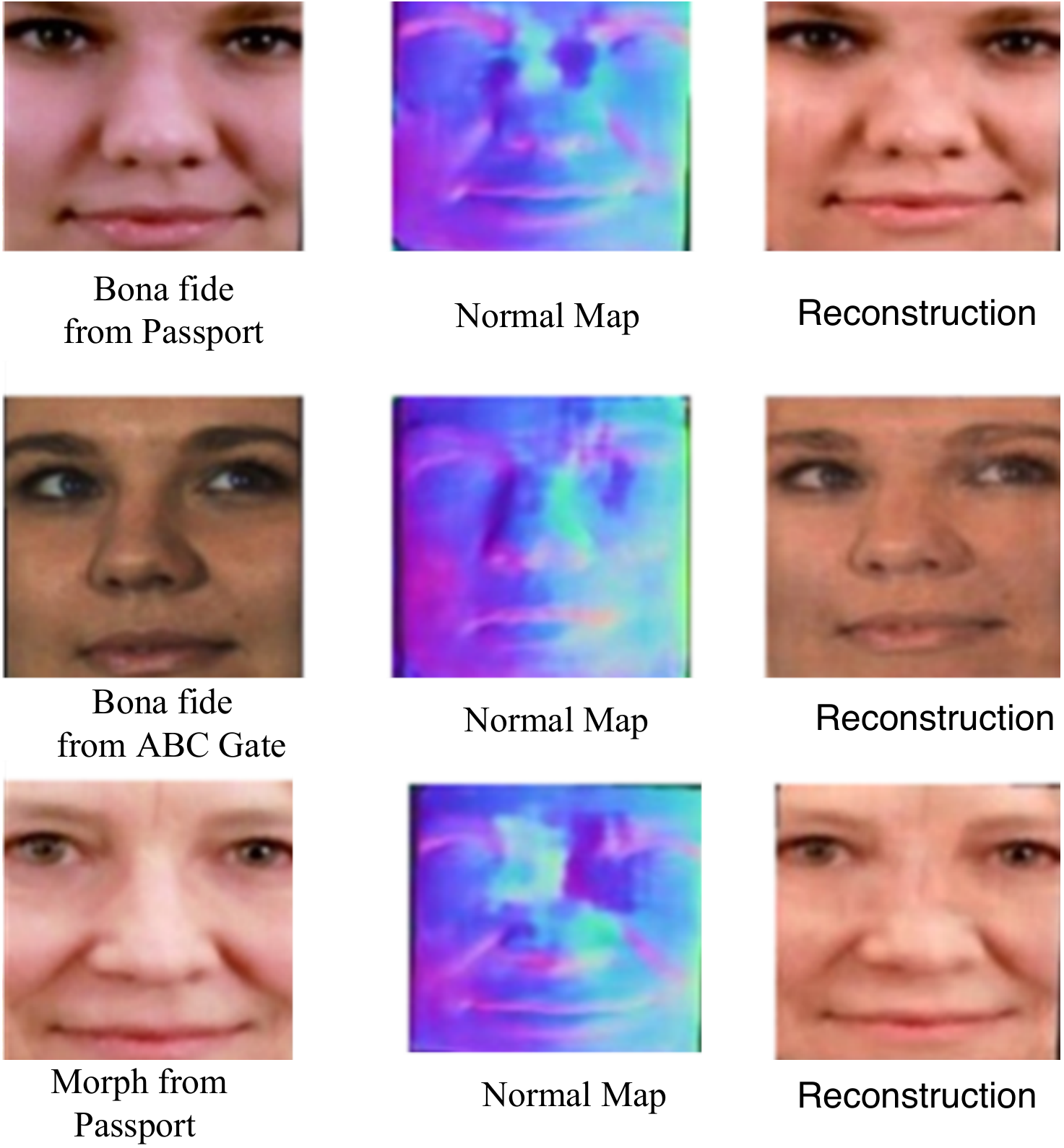} 
\caption{Image decomposed into Normal-Map and Reconstruction (Diffuse Reconstructed Image)}\label{fig:SFSNetResults}
\end{figure}
As it can be observed from Figure~\ref{fig:SFSNetResults}, the diffuse reconstructed image (pixel color differences are highlighted), and the normal-map (especially around the eye, and the nose regions) help to distinguish the bona fide and morph images, while in the non-decomposed domain they look quite similar. 

\subsubsection{Feature Extraction}
We extract the features as depicted in the  Figure~\ref{fig:FeatureExtractionAndClassification} within the proposed algorithm shown in Figure~\ref{fig:ProposedAlgorithm}. Owing to the robust nature of Alexnet~\cite{Alexnet12} in obtaining reliable features, we employ the Alexnet to derive features from the diffusely reconstructed image. Given that the image is diffuse, we assert that it is closer in feature space than input image $I$. We use $fc7$ layer of Alexnet for extracting features resulting in a feature vector of $4096$ elements on which we compute reconstruction-loss as L1-Loss. 
We compute a quantized normal map of 21-bits from the normal map, which is output by SfSNet~\cite{Soumyadip18} as quantization would result in the normal map being robust to small variations. This is followed by taking the simple difference as L1-Loss. 
\subsubsection{Feature Classification}
 Given the set of features, we train a linear SVM for diffuse reconstruction-loss, and normal-loss. The scores are fused by weighted fusion to generate the score for each camera. This is followed by a weighted sum-rule fusion of scores from each camera to achieve the final score, which can be used for the detection of morph, as shown in Figure~\ref{fig:ProposedAlgorithm}. The weights in both fusion steps are chosen based on a greedy search optimization algorithm~\cite{Raghavendra16}. The weights chosen for each camera are $0.7$ for the diffuse reconstructed image classifier and $0.3$ for the normal-map classifier. The weights chosen for the cameras are as follows, Camera1 $0.2$, Camera2 $0.3$, Camera3 $0.2$, and Camera4 $0.2$. 

\section{Experimental Setup \& Results}\label{resultssection}
In this section, we provide details on our database and the corresponding experimental protocols, following the results obtained. We report the performance of the proposed D-MAD  algorithm using the following  metrics defined in the International Standard ISO/IEC 30107-3~\cite{ISO2017} described as follows:
\begin{itemize}
    \item Attack Presentation Classification Error Rate (APCER), which is the mis-classification rate of morph attack presentations.
    \item Bona fide Presentation Classification Error Rate (BPCER), which is the mis-classification of bona fide presentation as morphs.
\end{itemize}
We also report Detection Equal Error Rate (D-EER $\%$) and detection error trade-off curves, to examine the rate of change of mis-classification errors.
\subsection{Morph ABC Database}
To simulate the operational scenario with attacks in the enrolment and trusted probe images from ABC gates, we created a new database in this work. We want to point out that in a realistic operational scenario, the digital image in the eMRTD may be bona fide or morphed. First, we generate a set of enrolment images for $39$ subjects captured in a realistic studio setting with multiple images using a Canon DSLR camera of $21$ megapixels. Secondly, we capture the face images of the same $39$ subjects in an ABC gate using a real-world equipment~\cite{Raghu15}. We employ a single image per subject from DSLR images as a bona fide passport image and treat the images, which were captured from the ABC Gate with four different cameras (one sample each) as bona fide probe images. Employing another session of DSLR images captured from the enrolment set up, we create a  morphed passport image dataset using the images from $39$ subjects and the approach and conditions mentioned in work by Raghavendra et al. in~\cite{Raghu17} specifically subjects not wearing glasses, and using the same gender, and ethnicity. The morphed images and bona fide images are printed and scanned using EPSON XP-860 Printer, and Scanner.

\begin{table}[h!]
\resizebox{1\linewidth}{!}
{
    \centering
    \begin{tabular}{|c|c|c|c|c|} 
    \hline
        ~    & \bf{Bona fide Passport} & \bf{Bona fide all} & \multicolumn{2}{|c|}{\bf{Morphed Passport}} \\ 
        ~  & ~ & \bf{ABC Gate Cameras} & \multicolumn{2}{|c|}{}    \\ \hline
       {Train}  &  {19} & {237} & \multicolumn{2}{|c|}{{52}} \\ \hline
       {Test} & {20} & {222} & \multicolumn{2}{|c|}{{38}} \\ \hline
        \multicolumn{5}{|c|}{\bf{Bona fide per ABC Gate Camera}} \\ \hline
         ~    & \bf{Camera1} & \bf{Camera2} & \bf{Camera3} & \bf{Camera4} \\ \hline \hline
       {Train}  &  {58} & {64} & {58} & {57} \\ \hline
       {Test} & {57} & {63} & {49} & {53} \\ \hline
    \end{tabular}}
    \caption{Dataset Details}
    \label{table:owndataset}
\end{table}

\begin{table*}[b]
\centering
  \begin{tabular}{|l|l|l|l|l|}
  \hline
    {\bf{Method}} & {\bf{Cam}} & \bf{EER} & \bf{BPCER20} & \bf{BPCER10} \\ \hline
 Signed  & 1 &   {43.7$\pm$0.2} & {90.5$\pm$0.3} & {83.4$\pm$0.3} \\  \cline{2-5}
 Distance~\cite{Naser18} & 2 &  {46.7$\pm$0.3} & {93.5$\pm$0.4} & {87.8$\pm$0.2} \\ \cline{2-5}
 & 3 &  {45.8$\pm$0.2} & {92.7$\pm$0.3} & {86.3$\pm$0.7}  \\ \cline{2-5} 
 & 4 &  {45.1$\pm$0.3} & {91.4$\pm$0.4} & {82.3$\pm$0.3} \\ \cline{2-5} 
 & {Fused} & {42.6$\pm$0.2} & {90.0$\pm$0.1} & {81.5$\pm$0.3} \\ \cline{2-5} \hline
LBP \&  & 1 &  {41.7$\pm$0.4} & {81.1$\pm$0.6} & {72.4$\pm$1.0} \\ \cline{2-5}
SVM~\cite{Ulrich18} & 2 & {42.7$\pm$0.5} & {82.5$\pm$0.6} & {73.5$\pm$0.8} \\ \cline{2-5}
& 3 &  {38.1$\pm$0.5} & {83.3$\pm$0.7} & {71.5$\pm$0.6} \\ \cline{2-5}
& 4 & {39.6$\pm$0.3} & {79.6$\pm$0.5} & {71.1$\pm$0.4} \\ \cline{2-5}
& {Fused} & {28.5$\pm$0.4} & {67.2$\pm$0.6} & {54.2$\pm$0.8} \\ \cline{2-5} \hline
Proposed  & 1 & {18.1$\pm$0.1} & {36.3$\pm$0.7} & {27.1$\pm$0.3} \\ \cline{2-5}
Method & 2 & {19.7$\pm$0.4} & {34.7$\pm$0.7} & {28.3$\pm$0.7} \\ \cline{2-5}
& 3 & {19.1$\pm$0.1} & {35.9$\pm$0.1} & {27.3$\pm$0.1} \\ \cline{2-5}
& 4 & {18.8$\pm$0.1} & {36.1$\pm$0.1} & {27.5$\pm$0.3}  \\ \cline{2-5}
& {\textbf{Fused}} & \textbf{8.6$\pm$0.1} & \textbf{13.9$\pm$0.4} & \textbf{7.5$\pm$0.1}  \\
\cline{2-5} \hline
\end{tabular}
\vspace{3mm}
\caption{Signed Distance by Damer et al. approach~\cite{Naser18} using author's implementation, LBP and SVM by Scherhag et. al~\cite{Ulrich18}, and the proposed method where BPCER20 is BPCER@APCER=5\%, and BPCER10 is BPCER@APCER=10\%}\label{Results}
 \end{table*}


Performance Protocol: In D-MAD, as we need two images for morph detection, we consider the bona fide passport images v/s bona fide gate images as the genuine class samples, and morph passport image v/s bona fide gate image as the attack class samples. 
We now go into details of the number of scores generated during training as follows: From the enrolment, we have 19 bona fide passport images, complemented with 52 morphed passport images. Further from Camera 1 in the ABC Gate we have 58 bona fide probe images, which results in $19\times58=1102$ genuine scores, and $52\times58=3016$ attack scores, Camera2 results in $19\times64=1216$ genuine scores, and $52\times64=3328$, Camera3 results in $19\times58=1102$ genuine scores, and $52\times58=3016$, and Camera4 results in $19\times57=1083$ genuine scores, and $52\times57=2964$ attack scores. The number of scores generated during testing is as follows: From the enrolment 20 bona fide passport images, complemented by 38 morphed passport images. From Camera1 in the ABC Gate we have 57 bona fide probe images, which results in $20\times57=1140$ genuine scores, and $38\times57=2166$ attack scores, Camera2 results in $20\times63=1260$ genuine scores, and $38\times63=2394$, Camera3 results in $20\times49=980$ genuine scores, and $38\times49=1862$, and Camera4 results in $20\times53=1060$ genuine scores, and $38\times53=2014$ attack scores. During fusion of scores of the four cameras, we reach 980 genuine scores, and 1862 attack scores as this are the minimum number of genuine and attack scores available in all four cameras during testing.

\begin{figure*}[htp]
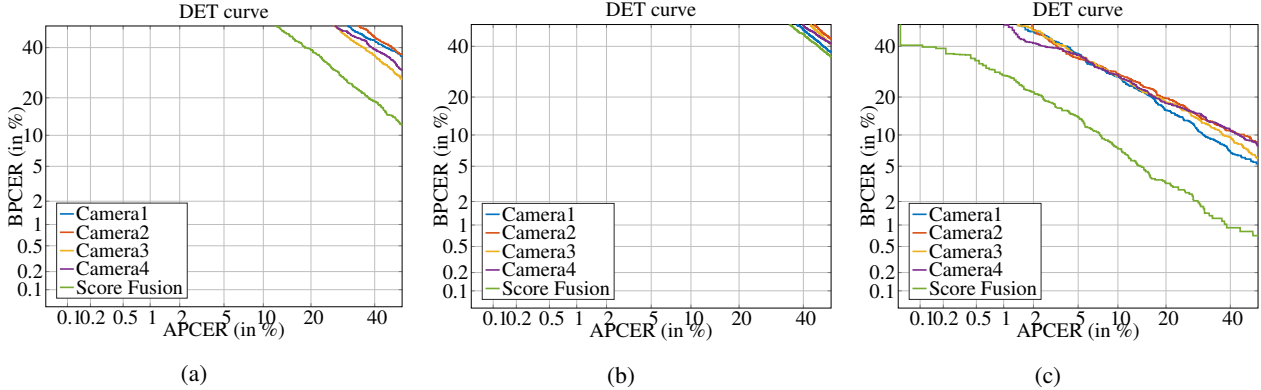

\begin{subfigure}{0.32\textwidth}
  \centering
\resizebox{1.0\linewidth}{!}{\input{UlrichDETCurve.tex}}
  \caption{}
  \label{fig:sub-first}
\end{subfigure}
\begin{subfigure}{0.32\textwidth}
  \centering
  \resizebox{1.0\linewidth}{!}{\input{naserdetcurveupdated.tex}}
  \caption{}
  \label{fig:sub-second}
 \end{subfigure}
\begin{subfigure}{0.32\textwidth}
  \centering
  \resizebox{1.0\linewidth}{!}{\input{DETCurveProposed.tex}}
  \caption{}
  \label{fig:sub-second}
 \end{subfigure}
 \caption{DET Curves for (a) LBP-SVM ~\cite{Ulrich18}, (b) Signed-Distance ~\cite{Naser18}, and (c) the proposed method. DET Curves are for Scores from Camera1, Camera2, Camera3, Camera4, and Weighted Sum-Rule Fusion of scores from these individual cameras.}
\label{fig:DETCurves}
\end{figure*}
\subsection{Analysis of Results}
Table~\ref{Results} presents the results of the proposed method and compares it with two state-of-the-art approaches including 
Landmark Shifts based Signed Distance proposed by Damer et al. ~\cite{Naser18} and Texture-Descriptors based LBP-SVM by Scherag et. al~\cite{Ulrich18}. As it can be noted from the Table~\ref{Results}, the proposed method outperforms existing SOTA, we achieve an EER of $8.6\pm0.1$ compared to best EER of SOTA of $28.5\pm0.4$. The results can also be seen in Figure~\ref{fig:DETCurves}, which presents the Detection Error Trade-off Curves, where it can be noted that fusion of scores leads to further improvement for the proposed algorithm compared to the SOTA. Despite outperforming the SOTA, we note that our proposed approach still has moderate deficiency from single cameras, as shown in Table~\ref{Results}. 
We make the following observations from the results:
\begin{itemize}
\item One can observe that in similar lighting capture environments, as shown in Figure~\ref{fig:MorphingExample} (row (a)), Image Subtraction based technique proposed by authors in~\cite{Ferrara18} performs well, and one can generalize this argument texture descriptor based method report by authors in~\cite{Ulrich18}. However, the same cannot be said for the technique proposed by authors in~\cite{Naser18} as landmark shifts could happen due to change in pose.
\item Figure~\ref{fig:MorphingExample} shows the degrading performance of the Image Subtraction based method proposed by authors in~\cite{Ferrara18} in (rows (b), and (c)) which have lighting changes and print-scan artifacts. The advantage of using features from a diffuse reconstructed image which contains lower-order lighting terms, and normal-map are shown in Figure~\ref{fig:SFSNetResults}.
\item The proposed method achieves the best D-EER compared to the existing SOTA mainly due to two factors, the use of a diffuse reconstructed image that removes the higher-order lighting components and leads to a linear light model without cast shadows as pointed out by Basri et. al~\cite{Basri03,Basri05}. The second factor is the use of normal-map, which on integration gives depth-map~\cite{Quéau2018}, and depth-map signifies the 3D shape of the bona fide sample. The 3D shape, and consequently normal-map of the bona fide sample, should be preserved across different cameras.
\end{itemize}

\section{Conclusion \& Future Work}\label{conclusionsection}
In this paper, we presented a novel and robust scheme to perform D-MAD in the presence of lighting, pose, and print-scan artifacts. We have constructed a new database reflecting the real-life border crossing scenario and have validated the results on our collected database. Our collected database models the real-life print-scan artifacts in the passport image and the use of camera images from the ABC gate. The proposed method outperforms the existing SOTA methods for D-MAD mainly due to the combined effect of pose normalization, use of a diffuse-reconstructed image, and normal map. In future works, the proposed algorithm shall be tested on a large scale database.

\section*{Acknowledgement}
This work was carried out under the funding of the Research Council of Norway under Grant No. IKTPLUSS 248030/O70. 
\balance
\bibliographystyle{ieee}
\bibliography{main}

\end{document}